\documentclass[twoside]{article}

\usepackage[accepted]{aistats2025}

\usepackage{microtype}
\usepackage{graphicx}
\usepackage{subfigure}
\usepackage{booktabs} 
\usepackage{multirow}

\usepackage{hyperref}

\usepackage{algorithm}
\usepackage{algorithmic}

\usepackage{amsmath}
\usepackage{amssymb}
\usepackage{mathtools}
\usepackage{amsthm}

\DeclareMathOperator*{\argmin}{arg\,min}

\theoremstyle{plain}
\newtheorem{proposition}{Proposition}[section]
\newtheorem{assumption}{Assumption}[section]

\usepackage[textsize=tiny]{todonotes}
\usepackage[round]{natbib}

\begin{document}

%
\runningtitle{MING: A Functional Approach to Learning Molecular Generative Models}

%

\twocolumn[

\aistatstitle{MING: A Functional Approach \\ to Learning Molecular Generative Models}

\aistatsauthor{ Van Khoa Nguyen \And Maciej Falkiewicz  \And  Giangiacomo Mercatali \And Alexandros Kalousis}

\aistatsaddress{ HES-SO Geneva \\ University of Geneva \And  HES-SO Geneva \\ University of Geneva \And HES-SO Geneva \And HES-SO Geneva} ]

\begin{abstract}
Traditional molecule generation methods often rely on sequence- or graph-based representations, which can limit their expressive power or require complex permutation-equivariant architectures. This paper introduces a novel paradigm for learning molecule generative models based on functional representations. Specifically, we propose Molecular Implicit Neural Generation (MING), a diffusion-based model that learns molecular distributions in the function space. Unlike standard diffusion processes in the data space, MING employs a novel functional denoising probabilistic process, which jointly denoises information in both the function's input and output spaces by leveraging an expectation-maximization procedure for latent implicit neural representations of data. This approach enables a simple yet effective model design that accurately captures underlying function distributions. Experimental results on molecule-related datasets demonstrate MING's superior performance and ability to generate plausible molecular samples, surpassing state-of-the-art data-space methods while offering a more streamlined architecture and significantly faster generation times. The code is available at \href{https://github.com/v18nguye/MING}{https://github.com/v18nguye/MING.}

\end{abstract}

\section{INTRODUCTION}

Finding novel, effective molecules has been a long-standing problem in drug design. A data-driven approach to solving this problem is to learn a generative model that can capture the underlying distributions of molecules and generate new compounds from the learned distributions. This approach requires an underlying structure of molecules, which typically takes sequence-based \citep{kusner2017grammar, gomez2018automatic} or graph-based \citep{shi2020graphaf, luo2021graphdf, jo2022score} representations. These representations have been adapted to many generative scenarios. However, these models either yield lackluster performance, require sophisticated equivariant architectures, or do not efficiently scale with molecule sizes.
\begin{figure}[t]
\centerline{
\includegraphics[width=.9\columnwidth]{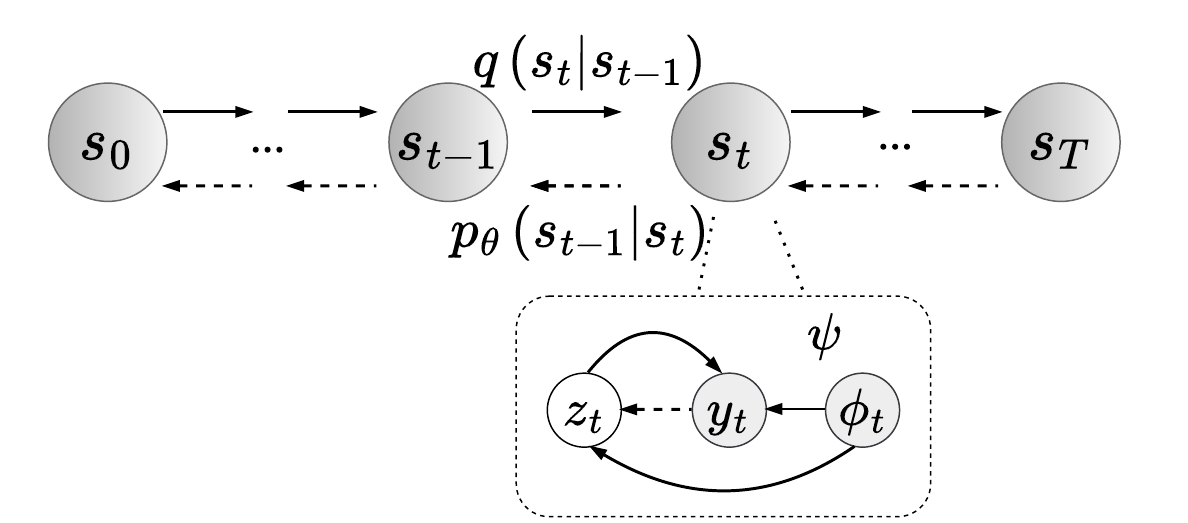}}
\caption{MING Graphical Model. (\textit{Top}) A forward process $q(s_{t} | s_{t-1})$ defines a noise schedule on function's output space. We introduce an INR-based denoising network $\theta$ that learns to estimate reverse diffusion processes on function space. (\textit{Bottom}) A noisy signal $y_t$ is a function of trainable latent $z_t$ and coordinate-wise inputs $\phi_t$. We parameterize the dependency by an another INR-based network $\psi$. The unobserved latent input $z_t$ can be obtained from the noisy output $y_t$ via gradient-descent optimization.}
\label{fig:graphical}
\end{figure}
Representing data as continuous, differentiable functions has become of surging interest for various learning problems \citep{park2019deepsdf, zhong2019reconstructing, mehta2021modulated, dupont2022data}. Intuitively, we represent data as functions that map from some coordinate inputs to signals at the corresponding locations relating to specific data points. A growing area of research explores using neural networks, known as implicit neural representations (INRs), to approximate these functions. These approaches offer some practical advantages for conventional data modalities. First, the network input is element-wise for each coordinate, which allows these networks to represent data at a much higher resolution without intensive memory usage. Second, INRs depend primarily on the choice of coordinate systems, thus a single architecture can be adapted to model various data modalities. Their applications have been explored for the representation tasks of many continuous signals on certain data domains such as images \citep{sitzmann2020implicit}, 3D shapes \citep{park2019deepsdf}, and manifolds \citep{grattarola2022generalised}.

Recently, an emerging research direction studies generative models over function space. The idea is to learn the distributions of functions that represent data on a continuous space. Several methods, including hyper-networks, \citep{dupont2021generative, du2021learning, koyuncu2023variational}, latent modulations \citep{dupont2022data}, diffusion probabilistic fields \citep{zhuang2023diffusion}, have addressed some challenging issues when learning functional generative models on regular domains such as images, and spheres. On the other hand, functional generative problems on irregular domains like graphs and manifolds still receive less attention. In a few works, \cite{elhag2024manifold, wang2024swallow} showed these tasks are feasible. They represent a data function on manifolds by concatenating its signal with coordinate inputs derived from the eigenvectors of graph Laplacians. The graph structure underlying these Laplacians is sampled from the manifold itself. They then apply the standard denoising diffusion probabilistic models \citep{ho2020denoising} to learn function distributions, and leverage a modality-agnostic transformer \citep{jaegle2021perceiver} as denoiser. While these approaches present some promising results, their training regimes rely on this particular architecture, which is computationally expensive and memory-inefficient for the continuous representation objectives. 

To the best of our knowledge, we propose the first generative model that learns molecule distributions and generates both bond- and atom- types by exploring their representations in function space. We achieve this by first introducing a new approach to parameterize both molecular-edge- and node-coordinate systems via graph-spectral embeddings. We then propose Molecular Implicit Neural Generation (MING), a novel generative model that applies denoising diffusion probabilistic models on the function space of molecules. Our method presents a new functional denoising objective, which we derive by utilizing an expectation-maximization process to latent implicit neural representations; we thus dub it as the expectation-maximization denoising process. 
By operating on function space, we are able to simplify the model design and introduce an INR-based architecture as a denoiser. Our model not only benefits from the advantages of INRs but also overcomes the constraints imposed by graph-permutation symmetry, resulting in a model complexity that is less dependent on molecular size. 
MING effectively captures molecule distributions by generating novel molecules with chemical and structural properties that are closer to test distributions than those generated by other generative models applied to data space. Additionally, MING significantly reduces the number of diffusion/sampling steps required.

\section{REPRESENTING MOLECULES AS FUNCTIONS}

In this section, we present a novel approach for representing functions of molecular graphs using implicit neural representations (INRs) on their irregular graph domain.

\subsection{Preliminaries}

\paragraph{Notation} We denote a molecule graph by $\mathcal{G} \triangleq \left(X, W\right)$, where $X$ and $W$ correspond to the matrices of node and edge features, respectively. For each graph $\mathcal{G}^i$, there exists a graph topology $\mathcal{T}^i \subset \mathbb{R}^d $ and its corresponding graph signals $\mathcal{Y}^i \subset \mathbb{R}^m $. The function representation of molecules is defined as:
\begin{align}
    f^i: \mathcal{T}^i \mapsto \mathcal{Y}^i
\end{align}
which maps from the function-input space, $ \mathcal{T}^i$, to the function-output space, $\mathcal{Y}^i$, where these output signals represent atom and bond types. One way to parameterize these functions is by using implicit neural representations (INRs). The approach employs a neural network ${\theta}$ as the function's parameters to model the relation between the input and output of the function.

\paragraph{Conditional INRs}
Standard INRs typically use a separate set of parameters $\theta_i$ for each function $f^i$, denoted as $f_{\theta_i}^i$, which is costly and inefficient for optimization.
To enable effective parameter sharing between molecules, we leverage conditional INR \citep{mehta2021modulated}, wherein a trainable latent input vector is conditioned on each function while sharing the model parameters $\theta$ across functions. We denote this latent input space by $\mathcal{Z} \subset \mathbb{R}^k$ and the conditional parameterization is given by:
\begin{align}
\label{eq:cinr}
    f_{\theta}^i: \mathcal{T}^i \times \mathcal{Z} \mapsto \mathcal{Y}^i
\end{align}
where $\mathcal{Z}$ is an unknown manifold, and $k \ll | \theta |$. In this way, we can use the shared network $\theta$ and only need to optimize a different latent input for each function, leading to more efficient optimization. 
\begin{figure*}[t]
\centerline{
\includegraphics[width=1.\textwidth]{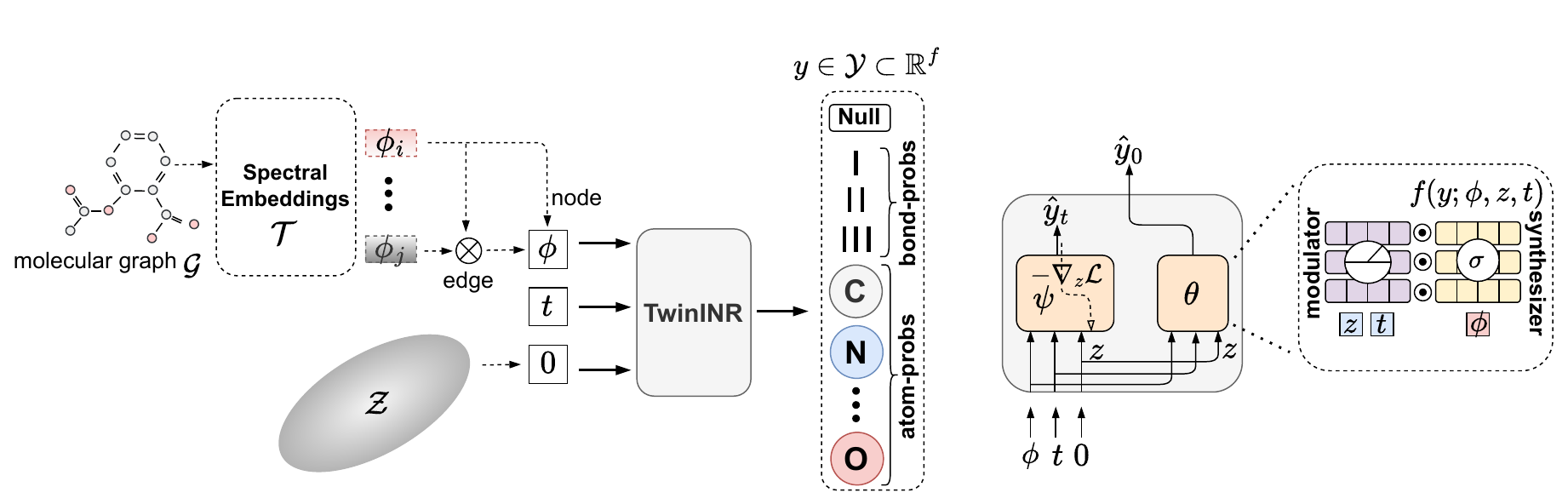}}
\caption{TwinINR Operational Flows. (\textit{Left}) TwinINR predicts a multi-dimension vector that encodes atom-type, bond-type, and a null token probability. While the node-coordinate system $\mathcal{T}$ bases on the eigenvector space of molecular graph Laplacian, we introduce the edge-coordinate system by taking the product of node-coordinate spaces $\mathcal{T}\times\mathcal{T}$. (\textit{Right}) The TwinINR's internal architecture is composed of two identical (twin) conditional INR networks $\psi$ and $\theta$. The latent network $\psi$ estimates the latent-input representation $z$ of the diffused signal $y_t$ at the diffusion step $t$ by maximizing the latent model likelihood. The denoising network $\theta$ predicts the clean signal $y_0$ from the noisy latent input $z$.  We adopt the latent-modulation strategy \citep{mehta2021modulated} for each network.}
\label{fig:arch}
\end{figure*}
\subsection{Parameterization on Irregular Domains}
\label{sec:param}
INRs require a coordinate system to define the function's input domain. Coordinate systems can be readily chosen for data on regular domains such as images or spheres using regular vertex coordinates. However, defining coordinate systems for data on irregular domains such as graph structures is challenging. 
In a relevant work, \cite{grattarola2022generalised} proposed using graph spectral embeddings as the coordinate system for INRs, computing the eigenvectors of graph Laplacians, and using them as node coordinate inputs. This work extends the approach to work on edge coordinate systems, which enables generative models to learn molecular bond types by INRs.

\paragraph{Node Input Coordinates}
For a molecular graph $\mathcal{G}$ and its weighted graph Laplacian $L_w = D - W$, where $D =  \text{diag}\left(\text{sum}(W_{:c})\right)$ with $c$ as a column index, we define the node coordinate system using the graph topology $\mathcal{T}$, which can be represented by the eigenvectors of $L_w$. A node input coordinate $\phi_{n} \in \mathcal{T}$ characterizes the topological structure surrounding it. The conditional-INR node mapping is given by $y = \theta(\phi_n, z)$, where $z\in \mathcal{Z}$ and $y$ encodes atom types. 

\paragraph{Edge Input Coordinates} 
Modeling the bond types of molecules using INRs requires the incorporation of edge input coordinate systems.  A straightforward approach is to use the pairwise product of node coordinates. Specifically, for a given edge between two nodes $\{i,j\}$, we define the edge input coordinate as $\phi_{e_{ij}} \triangleq \phi_{n_i} \odot \phi_{n_j}$, where $\odot$ is the Hadamard product. The conditional-INR edge mapping is then represented as $y = \theta(\phi_e, z)$, with $y$ encoding edge types, and $\phi_{n_i}, \phi_{n_j} \in \mathcal{T}$. 

\paragraph{Function Evaluations}
Since the explicit representation of functions is not accessible, we instead represent each function by a set of function evaluations $f \triangleq \{ s \triangleq (\phi, z, y) | \phi \in \mathcal{T} \left[ \times \mathcal{T}\right], z \in \mathcal{Z}, y \in \mathcal{Y}\}$, where each function evaluation $s$  is represented as a triplet that includes the function's input and output. Specifically, given a topology input $\phi$, we evaluate its corresponding function output $y$ and define the function evaluation as $s \triangleq (\phi, z, y)$. Here, $\phi$ can be a node- or edge-coordinate input, $y$ is the associated molecular signal type, and $z$ is a trainable latent input that facilitates parameter sharing between functions. We now present a novel denoising diffusion probabilistic model for molecule functions. 

\section{MOLECULAR IMPLICIT NEURAL GENERATION}
In this section, we introduce novel functional diffusion posteriors and present an expectation-maximization denoising process as the core optimization algorithm for functional denoising diffusion probabilistic models. Building on this foundation, we propose TwinINR, a novel INR-based architecture specifically designed for denoising tasks. Our generative model is tailored for molecular generation problems, a framework we refer to as Molecular Implicit Neural Generation (MING).

\subsection{Functional Probabilistic Posteriors}
The goal is to train a probabilistic denoising model that maximizes the model likelihood $p_{\theta}\left(s_0\right)=\int p_{\theta}\left(s_{0:T}\right)ds_{1:T}$ under the reverse diffusion process $p_{\theta}\left(s_{0:T}\right)$. Following the relevant works \citep{sohl2015deep, ho2020denoising}, we show that this objective is equivalent to minimizing the Kullback-Leibler divergence between the conditional forward posterior $q(s_{t-1}|s_{t}, s_{0})$ and the denoising model posterior $p_{\theta}(s_{t-1}|s_{t})$, where $s_t \triangleq (\phi_t, z_t, y_t)$ is a function evaluation on noisy molecule $\mathcal{G}_t$.

A key challenge is to define noise kernels on the forward diffusion process and estimate data scores for the reverse generative diffusion process within the function space. Since the function maps from the input domain to the output-signal space, we thus simplify the problem by only diffusing noises on the output-signal space to get noisy functions. Concretely, given a molecular graph $\mathcal{G}_0$, which function evaluation on a topology $\phi_0$ is $s_0 \triangleq \left(\phi_0, z_0, y_0 \right)$, we make the mild assumption:
\begin{assumption}
   The topological information $\phi_0$ remains subtly variant throughout the diffusion process. However, the clean signal output, $y_0$, which resides on the topological location, undergoes diffusion and converges to a normal prior. This process generates a series of noisy molecular function evaluations $s_t \triangleq \left(\phi_0, z_t, y_t\right)$, where $z_t$ is the noisy latent input evolving in response to the diffused signal $y_t$.
\end{assumption}

\subsubsection{Conditional Forward Posterior}
As mentioned above, we apply noise exclusively to the function's output component while keeping the topology-input part unchanged. As a result, the posterior satisfies $\phi_{t} \simeq \phi_{0}$ throughout the diffusion process. By expressing the function evaluation in terms of the triplet-based representation, the posterior becomes:
\begin{align}
    q(s_{t-1}|s_{t}, s_{0}) &= q (\phi_{t-1}, z_{t-1}, y_{t-1} | \phi_{t}, z_{t}, y_{t}, \phi_{0}, z_{0}, y_{0}) \nonumber \\
    &= q (z_{t-1}, y_{t-1} | z_{t}, y_{t}, \phi_{0}, z_{0}, y_{0}) \nonumber \\
    &= q (z_{t-1} | y_{t-1}, \phi_{0}) q(y_{t-1} | y_{t}, y_{0})
\end{align}
The detailed derivation is in Appendix \ref{proof:cfp}. In the third equality, we apply Bayes' rule. For the first term, we assume temporal conditional independence between the current latent input $z_{t-1}$ and the function-input $z_{t}$ and output variables $y_{t}$ at other diffusion steps. We formalize this temporal conditional independence by another mild assumption below:  
\begin{assumption}
    The latent input variable \(z_t\) can be reconstructed from its corresponding noisy signal output \(y_t\) at the same diffusion step and along with the fixed topology input \(\phi_0\), which is independent of the function's input and output variables evaluated at other diffusion steps.
\end{assumption}
For the second term, we employ the standard denoising diffusion probabilistic process \citep{ho2020denoising} over the signal space $\mathcal{Y}$, which defines its noise kernel and conditional forward posterior as follows:
\begin{align}
\label{eq:q}
    q(y_t | y_0) &= \mathcal{N}(\mu_{y_t| y_0}, \sigma_{y_t| y_0}^2 \mathbf{I}) \nonumber \\
    & \mu_{y_t| y_0} = c_0(t) y_0 \nonumber \\
    & \sigma_{y_t| y_0} = c_1(t) \nonumber \\
    q(y_{t-1} | y_t, y_0)  &= \mathcal{N}(\mu_{y_{t-1}|y_t, y_0}, \sigma_{y_{t-1}|y_t, y_0}^2 \mathbf{I}) \nonumber \\
    & \mu_{y_{t-1}|y_t, y_0} = c_2(t) y_0 + c_3(t) y_t \nonumber \\
    & \sigma_{y_{t-1}|y_t, y_0} = c_4(t)
\end{align}
where $c_i(t)$s are the diffusion coefficients depending on the diffusion step $t$, as introduced in \cite{ho2020denoising}.

\subsubsection{Denoising Model Posterior}
We substitute the triplet-based representation of function evaluation into the denoising model posterior:
\begin{align}
p_{\theta}(s_{t-1} | s_{t}) &= p_{\theta}(z_{t-1}, \phi_{t-1}, y_{t-1} | z_{t}, \phi_{t}, y_{t}) \nonumber \\
&= p_{\theta}(z_{t-1}, y_{t-1} | z_{t}, \phi_{0}, y_{t}) \nonumber\\
&= p_{\theta}(z_{t-1} | \phi_{0}, y_{t-1}) p_{\theta}(y_{t-1} | z_{t}, \phi_0, y_t)
\end{align}
where we apply the topology-input invariance assumption throughout diffusion process to the first equality and the latent temporal conditional independence assumption to the second equality. We visualize the MING's graphical model in Figure \ref{fig:graphical}.

\subsection{Functional Denoising Probabilistic Models}

The objective is to minimize the Kullback-Leibler divergence between the functional posteriors at every diffusion step, $\mathcal{L}_{t-1}=D_{KL}(q(s_{t-1} | s_{t}, s_{0}) \| p_{\theta}(s_{t-1} | s_{t}))$. We prove that the divergence can be simplified into two KL terms optimized directly on the signal and latent spaces by the following proposition:
\begin{proposition}
\label{prop:L}
    For a triplet-based representation of molecule function evaluation $s_{t} \triangleq (\phi_{t}, z_{t}, y_{t})$, by assuming the topological invariance $\phi_{t} \simeq \phi_{0}$ and the temporal conditional independence of latent posterior $q(z_{t}|\phi_0, y_{t})$, the KL divergence between two functional posteriors $D_{KL}(q(s_{t-1} | s_{t}, s_{0}) \| p_{\theta}(s_{t-1} | s_{t}))$ splits into the sum of two KL divergences on the spaces $\mathcal{Z}$ and $\mathcal{Y}$:
\end{proposition}

\scalebox{.9}{
\begin{minipage}{\linewidth}
\begin{align}
\label{eq:obj1}
\mathcal{L}_{t-1} &= \mathcal{L}_{t-1}^\mathcal{Z} + \mathcal{L}_{t-1}^\mathcal{Y}\nonumber \\
=\mathbb{E}_{q(y_{t-1} | y_t, y_0)}&\left(D_{KL}\left(q(z_{t-1}|y_{t-1}, \phi_0) \| p_{\theta} (z_{t-1} | y_{t-1}, \phi_0)\right)\right) \nonumber \\
&+ D_{KL} \left(q(y_{t-1}| y_t, y_0) \| p_{\theta}(y_{t-1}|z_t, \phi_0, y_t)\right)
\end{align}
\end{minipage}
}

We give the proof in Appendix \ref{proof:p1}. The first term, $\mathcal{L}_{t-1}^\mathcal{Z}$, represents the KL divergence computed over the latent-input space $\mathcal{Z}$, which then takes the expectation w.r.t the conditional forward posterior on the signal space $\mathcal{Y}$. The second term, $\mathcal{L}_{t-1}^\mathcal{Y}$, quantifies the KL divergence between two posteriors on $\mathcal{Y}$. Below, we will analyze the KL minimization problem in each space.

\subsubsection{Gradient Origin $\mathcal{Z}$-based Optimization} 
We now analyse the KL term $\mathcal{L}_{t-1}^\mathcal{Z}$ on the latent-input space $\mathcal{Z}$. However, there are no closed-form solutions for the posteriors of latent variable $z_{t-1}$. We thus approach this task as a likelihood-maximization problem given a noisy observed signal $y_{t-1}$, an observed topology $\phi_0$, and an unobserved latent input $z_{t-1}$. 
To achieve this, we introduce a latent model $p_{\psi}(z_{t-1},  y_{t-1}, \phi_0)$ parameterized by an auxiliary INR network $\psi$ that learns the mapping between noisy function's output and inputs, ${\psi}: \phi_0 \times z_{t-1} \mapsto y_{t-1}$. We can show that the KL term $\mathcal{L}_{t-1}^\mathcal{Z}$ converges to zeros by the next proposition:
\begin{proposition}
    \label{prop:Z}
    For the likelihood optimization problem involving the latent model $p_{\psi}(z_{t-1}, y_{t-1}, \phi_0)$ shared across both the denoising-model, $p_{\theta} (z_{t-1} | y_{t-1}, \phi_0)$, and conditional forward, $q(z_{t-1}|y_{t-1}, \phi_0)$, posteriors on $\mathcal{Z}$, the KL divergence $\mathcal{L}_{t-1}^\mathcal{Z}$ between the two posteriors converges to zeros as they have the same empirical distribution at each reverse diffusion step, or $\mathcal{L}_{t-1}^\mathcal{Z} \rightarrow 0$. 
\end{proposition}
\textit{Proof:} We remark that these two latent posteriors are conditioned on the same factors, namely $\phi_0$ and $y_{t-1} \sim q(y_{t-1} | y_0, y_t)$. By employing the same latent model $p_{\psi}(z_{t-1}, y_{t-1}, \phi_0)$ applied to both the denoising model and conditional forward posteriors, we have that these posteriors exhibit similar latent distributions in $\mathcal{Z}$. A more detailed proof is in Appendix \ref{proof:p2}.

However, we still need to estimate the noisy-latent input $z_t$ for the denoising process. Inspired by similar work \citep{bond-taylor2021gradient}, we first initialize the latent-input variable $z_t$ at the origin of $\mathcal{Z}$, which helps to reduce input variance and thus to stabilize the optimization process. We then iteratively optimize $z_t$ and $\psi$ by gradient descent to maximize the latent model log-likelihood $p_{\psi}(z_{t}, y_{t}, \phi_0)$, which is equivalent to minimizing the reconstruction loss in the signal space $\mathcal{Y}$:

\scalebox{.83}{
\begin{minipage}{\linewidth}
\begin{align}
    \label{eq:z}
    \mathcal{L}_{out}(\psi)= \int \mathcal{L}\left (y_t, {\psi} \left(\phi_0, \underbrace{- \nabla_{z_t} \int \mathcal{L}\left(y_t, {\psi}(\phi_0, z_t)\right)d\phi_0}_{\mathcal{L}_{in}(z_t)}\right) \right)d\phi_0
\end{align}
\end{minipage}
}

where $\mathcal{L}$ denotes the mean-squared error loss, and both integrals are evaluated over the topological space $\mathcal{T}$. 
The main objective consists of two loss components. The first term, referred to as the inner loss $\mathcal{L}_{in}(z_t)$, involves fixing the network parameters $\psi$ while optimizing $z_t$, which is initially set at the origin. The second term, known as the outer loss $\mathcal{L}_{out}(\psi)$, involves optimizing the network parameters $\psi$ using the $z_t$ derived from the inner-optimization step. These two steps are analogous to performing the expectation-maximization (E-M) optimization over the latent model $p_{\psi}(z_{t}, y_{t}, \phi_0)$. Below, we introduce a novel training objective for diffusion models applied to the function space of molecules. 

\subsubsection{E-M Denoising Process} 

\begin{algorithm}[t]
   \caption{Molecular Implicit Neural Generation.}
   \label{alg:fpdm}
\begin{algorithmic}
    \STATE {\bfseries Input}: denoiser $\theta$,  latent model $\psi$, $T$ diffusion steps, training set $\mathcal{D}_{train}$. \\
    /* \textit{Training} */
    \STATE  {\bfseries for} $(\phi_{0} = \phi_{n}[,\phi_{e}], y_0) \in \mathcal{T}^{i} \left[ \times \mathcal{T}^{i}\right] \times \mathcal{Y}^{i} \sim \mathcal{D}_{train}$:
     \STATE \hspace{10pt} $t \sim \mathcal{U}_{[0, T]}$
    \STATE \hspace{10pt} $y_t \sim \mathcal{N}\left(\mu_{y_t | y_0}, \sigma_{y_t | y_0}^2 \mathbf{I}\right)$ 
    \STATE \hspace{10pt} $z_t \leftarrow 0$
    \STATE \hspace{10pt}  {\bfseries for} \# iterations:
    \STATE \hspace{20pt} $\hat{y}_t \leftarrow {\psi}\left(z_t, \phi_0 \right) $
    \STATE \hspace{20pt} $\mathcal{L}_{e} \leftarrow \| \hat{y}_t - y_t\|^2 $
    \STATE \hspace{20pt} $z_t \leftarrow \text{gradient\_descent} \left(\nabla_{z_t} \mathcal{L}_{e}\right)$ \COMMENT{E-step}
    \STATE \hspace{10pt} $\hat{y}_0, \hat{y}_t \leftarrow {\theta}\left(z_t, \phi_0 \right), {\psi}\left(z_t, \phi_0 \right)$
    \STATE \hspace{10pt} $\mathcal{L}_m \leftarrow  \|\hat{y}_t - y_t\|^2 + \|\hat{y}_0 - y_0\|^2$
    \STATE \hspace{10pt} $\theta, \psi \leftarrow \text{adam} \left(\nabla_{\theta, \psi} \mathcal{L}_m \right)$ \COMMENT{M-step} \\
     /*  \textit{Sampling} */
    \STATE $y_T \sim \mathcal{N}\left(0, I\right)$
    \STATE $\phi_{0} = \phi_{n}[,\phi_{e}] \in \mathcal{T}^{i} \left[ \times \mathcal{T}^{i}\right] \sim \mathcal{D}_{train}$
    \STATE  {\bfseries for}  t $\rightarrow$ $[T, 0)$:\\
    \STATE \hspace{10pt} $z_t \leftarrow 0$
    \STATE \hspace{10pt}  {\bfseries for} \# iterations:
    \STATE \hspace{20pt} $\hat{y}_t \leftarrow {\psi}\left(z_t, \phi_0 \right) $
    \STATE \hspace{20pt} $z_t \leftarrow \text{gradient\_descent}\left(\nabla_{z_t} \|\hat{y}_t - y_t\|^2\right)$
    \STATE \hspace{10pt} $y_{t-1} \leftarrow \mathcal{N}(\mu_{y_{t-1}|z_t, \phi_0, y_t}, \sigma_{y_{t-1}|z_t, \phi_0, y_t}^2 \mathbf{I})$ 
    \STATE {\bfseries return}: $y_0$
\end{algorithmic}
\end{algorithm}

By substituting the previous result, $\mathcal{L}_{t-1}^\mathcal{Z} \rightarrow 0$, into the KL divergence in Equation \ref{eq:obj1}, we now only need to minimize the KL divergence term on $\mathcal{Y}$: 
\scalebox{.95}{
\begin{minipage}{\linewidth}
\begin{align}
   \mathcal{L}_{t-1} = \mathcal{L}_{t-1}^\mathcal{Y} = D_{KL} \left(q(y_{t-1}| y_t, y_0) \| p_{\theta}(y_{t-1}|z_t, \phi_0, y_t)\right)
\end{align}
\end{minipage}
}
The conditional forward posterior $q(y_{t-1}| y_t, y_0)$ is given in  Equation \ref{eq:q}, which we utilize to parameterize the denoising model posterior on the signal space:
\begin{align}
    p_{\theta}(y_{t-1} &| z_t, \phi_0, y_t) = \mathcal{N}(\mu_{y_{t-1}|z_t, \phi_0, y_t}, \sigma_{y_{t-1}|z_t, \phi_0, y_t}^2) \nonumber \\
    &\mu_{y_{t-1}|z_t, \phi_0, y_t} = c_2(t) \hat{y}_0 + c_3(t) y_t \nonumber \\             
    &\qquad\qquad\quad\,\, = c_2(t){\theta}(\phi_0, z_t) + c_3(t)y_t \nonumber \\
    & \sigma_{y_{t-1}|z_t, \phi_0, y_t} = c_4(t)
\end{align}
Here, the INR denoiser $\theta$ predicts the clean signal $y_0$ from the noisy latent input $z_t$ and the coordinate input $\phi_0$. Similar to the approach in \cite{ho2020denoising}, we can derive the denoising objective on $\mathcal{Y}$ as $\mathcal{L}_{t-1} = \| {\theta}(\phi_0, z_t) - y_0 \| ^ 2$, with $z_t$ optimized from the E-M process on the latent model $p_{\psi}(z_{t}, y_{t}, \phi_0)$ in Equation \ref{eq:z}. We thus refer to the overall optimization as the expectation-maximization denoising process. At every diffusion step, we solve two sub-optimization problems:

\paragraph{Expectation Step} We optimize the noisy latent input value $z_t$, initially set to the origin, by minimizing the inner loss $\mathcal{L}_{in}(z_t)$ with gradient descent, the expectation loss $\mathcal{L}_e$ is equal to the inner loss $\mathcal{L}_e = \mathcal{L}_{in}(z_t)$.

\paragraph{Maximization Step} We jointly optimize the parameters of denoiser $\theta$ using the denoising objective loss $\mathcal{L}_{t-1}$, while simultaneously optimizing the latent model parameters $\psi$ to maximize the latent model likelihood, given the obtained representation $z_t$, through $\mathcal{L}_{out}(\psi)$; we get the maximization loss as $\mathcal{L}_{m}=\mathcal{L}_{t-1}+\mathcal{L}_{out}(\psi)$. 

\subsection{Training and Sampling}

 In training, we use the node coordinates $\phi_0 = \phi_n$, edge coordinates $\phi_0 =\phi_e$, and their corresponding signals $y_0$ of molecular graphs $\mathcal{G}^i \triangleq \left(\mathcal{T}^i, \mathcal{Y}^i\right)$, which we sample from a training set $\mathcal{D}_{train}$.
 At each diffusion step, with $t$ uniformly distributed, $ \mathcal{U}_{[0, T]}$, we initialize the latent input variable $z_t$ with a zero-valued vector and optimize the expectation maximization denoising objective. 

To generate novel molecules, we first sample node and edge coordinate inputs from the training set $\phi_{0} = \phi_{n}[,\phi_{e}] \in \mathcal{T}^{i} \left[ \times \mathcal{T}^{i}\right] \sim \mathcal{D}_{train}$, and then initialize their signals from the normal prior $y_T \sim \mathcal{N}\left(0, I\right)$. We apply the reverse functional diffusion process, in which we only optimize the expectation objective at each reverse step to get the final generated signals, $y_{t=0}$. We summarize MING in Algorithm \ref{alg:fpdm}.

\subsection{TwinINR as Denoiser Architecture}

We design the denoising architecture using implicit-neural-representations (INRs), which consist of a stack of MLP layers combined with nonlinear activation functions. The architecture includes two identical INRs, namely TwinINR. The first module is the latent model network $\mathcal{\psi}$ that represents the noisy molecule function evaluations $s_t \triangleq (\phi_0, z_t, y_t)$. The second module often refers as the denoising network $\theta$ that predicts the clean signal $y_0$ from the perturbed latent input $z_t$ and the coordinate input $\phi_0$. We adopt the modulated conditional INR architecture \citep{mehta2021modulated} that allows for the efficient parameter sharing between molecule samples. Moreover, we extend the existing architecture to work on arbitrary domains, moving beyond regular vertex spaces like image or sphere domains, by using coordinates computed from graph spectral embeddings. 

The two components $\psi$ and $\theta$ share the same architecture design, with each containing a synthesis network and a modulation one. The synthesis network generates the signal output $y$ from the corresponding coordinate $\phi_0$, each layer is an MLP equipped with a nonlinear activation function:
\begin{align}
    h_i = \alpha_i\odot\sigma(w_i h_{i-1} + b_i)
\end{align}
where $h_i$ is the hidden feature at the $i$-th layer and $h_0 = \phi_0$,  $w_i$ and $b_i$ are the learnable weights and biases for the $i$-th MLP layer, $\sigma$ is a nonlinear activation, and $\alpha_i$ is the modulation vector that modulates the hidden feature output using the pointwise multiplication. We model the hidden modulation features using the modulation network, which is an another MLP stack with the activation ReLU:
\begin{align}
    \alpha_0 & = \text{ReLU}\left(w_{0}^{'}z + b_{0}^{'}\right) \nonumber \\
    \alpha_i &= \text{ReLU}\left(w_{i}^{'}[\alpha_{i-1} \; z]^{T} + b_{i}^{'}\right)
\end{align}
where $w_{i}^{'}$ and $b_{i}^{'}$ represent the weights and biases. Note that the latent input $z$  is concatenated at each layer of the modulation network. We illustrate TwinINR in Figure \ref{fig:arch}.

\section{EXPERIMENTS}

\begin{table*}[t]
\caption{Comparisons of models trained on the graph representation of molecules. The baseline results are sourced from GDSS \citep{jo2022score}, and GraphArm \citep{kong2023autoregressive}. Hyphen (-) denotes unreproducible results. The detailed results with standard deviations are in Appendix \ref{sec:exp}. }
\label{tab:graph}
\vskip 0.15in
\begin{center}
\begin{small}
\begin{tabular}{ll ccc ccc}
\toprule
\multirow{2}{*}{Method} & & \multicolumn{3}{c}{\textsc{QM9}} & \multicolumn{3}{c}{\textsc{ZINC250k}}\\
\cmidrule(lr){3-5} \cmidrule(l){6-8}
 & & Val. $ \uparrow$ &  NSPDK $\downarrow$ & FCD $\downarrow$ &  Val. $ \uparrow$ &  NSPDK $\downarrow$ & FCD $\downarrow$  \\
\midrule
\multirow{4}{*}{\textit{Graph}} & GraphAF \tiny{\citep{shi2020graphaf}} & 74.43 & 0.021  & 5.63 & 68.47 & 0.044 & 16.02 \\
& GraphDF \tiny{\citep{luo2021graphdf}} & 93.88 & 0.064  & 10.93 & 90.61 & 0.177 & 33.55 \\
& GDSS \tiny{\citep{jo2022score}} &  95.72 &  0.003 & 2.90 & 97.01 & 0.019 & 14.66\\
& DiGress \tiny{\citep{vignac2023digress}} &  99.00 &  \textbf{0.0005} & \textbf{0.36} & 91.02 & 0.082 & 23.06\\
& GraphArm \tiny{\citep{kong2023autoregressive}} & 90.25 & 0.002 &  1.22 & 88.23 & 0.055 & 16.26\\
& CatFlow \tiny{\citep{eijkelboom2024variational}} & \textbf{99.81} & - &  0.44 & \textbf{99.21} & - & 13.21\\
\midrule
\textit{Function} & MING & 98.23 & 0.002 & 1.17 & 97.19 & \textbf{0.005} & \textbf{4.80} \\
\bottomrule
\end{tabular}
\end{small}
\end{center}
\vskip -0.1in
\end{table*}

\begin{table}[t]
\caption{Comparisons with models trained on the sequence-based representation of molecules in QM9. The baseline results are taken from \cite{de2018molgan}.}
\label{tab:seq}
\vskip 0.15in
\begin{center}
\begin{small}
\begin{tabular}{ll ccc}
\toprule
 Method &  & Val.$\uparrow$  & Uni.$\uparrow$ &  Nov.$\uparrow$ \\
\midrule
\multirow{2}{*}{\textit{Sequence}}  & Character-VAE & 10.3 & 67.5 & \textbf{90.0} \\
& Grammar-VAE & 60.2 & 9.3 & 80.9 \\
\midrule
\textit{Function} & MING & \textbf{98.23} & \textbf{96.83} & 71.59 \\
\bottomrule
\end{tabular}
\end{small}
\end{center}
\vskip -0.1in
\end{table}

\subsection{Experimental setup}
\paragraph{Datasets}  We validate MING with respect to the ability to learn complex molecule structures in the function space. We benchmark on two standard and one large-scale molecule datasets. QM9 \citep{ramakrishnan2014quantum} contains an enumeration of small valid chemical molecules of 9 atoms, which are composed of 4 atom types: fluorine, nitrogen, oxygen, and carbon. Next, ZINC250k \citep{irwin2012zinc} is a larger molecule dataset that contains molecules up to the size of 38 atoms. The dataset is composed of 9 different atom types, exposing more diverse structural and chemical molecular properties. Finally, MOSES \citep{polykovskiy2020molecular} is the largest dataset that, to our best knowledge, has ever been experimented with for generative modeling tasks. The dataset contains up to $1.9$ Million molecules, whose size can reach 27 atoms per molecule. Following \cite{jo2022score}, we kekulize molecules by RDKit \citep{landrum2016rdkit} and remove hydrogen atoms.

\paragraph{Metrics} We measure molecule-generation statistics in terms of validity without correction (Val.), which is the percentage of chemical-valid molecules without valency post hoc correction. Among the valid generated molecules, we compute uniqueness (Uni.) and novelty (Nov.) scores, which measure the percentage of uniquely generated and novel molecules compared to training sets, respectively. In addition to the generation statistics, we quantify how well the molecule distributions generated align with the test distributions by computing two more important metrics, namely the Fréchet ChemNet Distance (FCD) \citep{preuer2018frechet} and the Neighborhood Subgraph Pairwise Distance Kernel (NSPDK) \citep{costa2010fast}. While FCD measures the chemical distance between two relevant distributions, NSPDK compares their underlying graph structure similarity. These two metrics highlight the quality of generated molecules more precisely by taking into account the molecule's structural and chemical properties rather than only their generation statistics. 

\paragraph{Implementations} 
We adhere to the evaluation protocol outlined in GDSS \citep{jo2022score}, in which we utilize the same data splits. To assess the impact of hyperparameter settings, we ablate different hidden feature dimensions $\{64, 128, 256\}$, latent z dimensions $k \in \{8, 16, 32\}$, coordinate input dimensions $d \in \{10, 20, 30, 37\}$ on ZINC250k, $d \in \{5, 7, 9\}$ on QM9 and $d \in \{20, 25\}$ on MOSES. We set the inner optimization iterations to three for all experiments. We observe that MING performs competitively with only 100 diffusion steps on QM9, 30 diffusion steps on ZINC250k, and 100 diffusion steps on MOSES.  The main results report the mean of three samplings, each consisting of 10000 sampled molecules for ZINC25k, QM9, and 25000 sampled molecules for MOSES. The hyperparameter settings are provided in Appendix \ref{sec:exp}.

\begin{table}[t]
\caption{Comparison of models trained on the MOSES dataset. Hyphen (-) indicates unreproducible results.}
\label{tab:moses}
\vskip 0.15in
\begin{center}
\begin{small}
\begin{tabular}{ll cccc}
\toprule
 Method &  & Val.$\uparrow$  & Uni.$\uparrow$ &  Nov.$\uparrow$ &  FCD $\downarrow$\\
\midrule
\multirow{1}{*}{\textit{Graph}}  & DiGress & 85.7 & \textbf{100} & 95.0 & - \\
\midrule
\textit{Function} & MING & \textbf{97.23} & 99.88 & \textbf{100} & 15.03\\
\bottomrule
\end{tabular}
\end{small}
\end{center}
\vskip -0.1in
\end{table}

\subsection{Baselines}
We benchmark MING against two types of molecular generative models, which utilize either the sequence-based or graph-based representation of molecules. Specifically, we compare Grammar-VAE \citep{kusner2017grammar} and Character-VAE \citep{gomez2018automatic}; they apply variational autoencoders to SMILES-based representations. In graph space, we compare with GraphAF \citep{shi2020graphaf}, a continuous autoregressive flow model, GraphDF \citep{luo2021graphdf}, a flow model for the discrete setting, GDSS \citep{jo2022score}, a continuous graph score-based framework, DiGress \citep{vignac2023digress}, a discrete diffusion model on graphs, GraphArm \citep{kong2023autoregressive}, an autoregressive diffusion model, and CatFlow \citep{eijkelboom2024variational}, a recent variational flow matching framework for graphs. 

\subsection{Results}

\begin{figure}[th]
\centerline{
\includegraphics[width=.8\columnwidth]{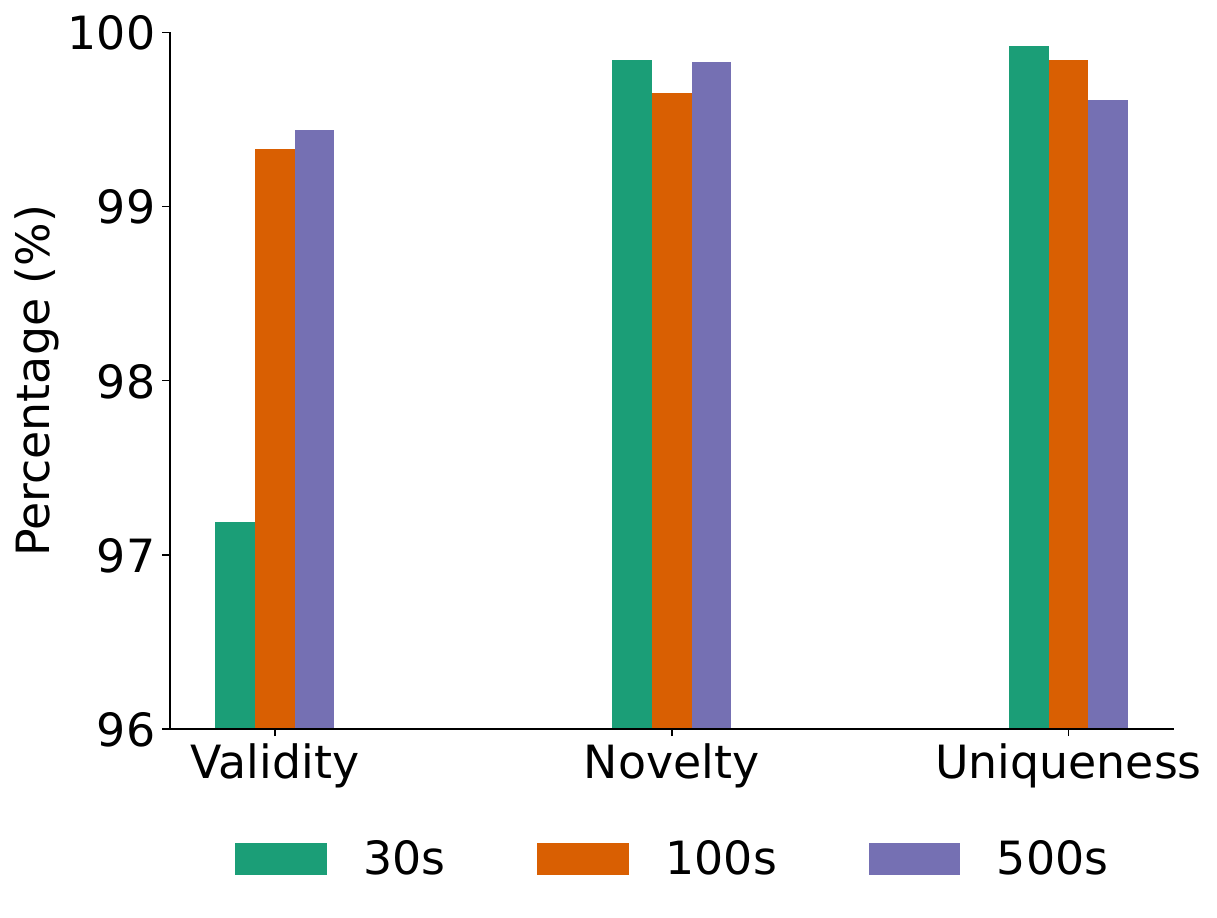}}
\caption{Validity, uniqueness, and novelty results across different number of diffusion time steps $T$ on ZINC250k}
\label{fig:dstep}
\end{figure}

\paragraph{Graph Generation Statistics } We evaluate the graph generation statistics based on validity, uniqueness, and novelty. Table \ref{tab:graph} suggests that MING demonstrates competitive performance in sampling chemical valid molecules compared with the graph-based baselines. We attribute this result to MING, which removes the graph-permutation constraint by generating signals directly on the topology inputs of data, a process typically employed in function-based generative frameworks. In contrast, some baselines struggle to learn the graph permutation symmetry by either preassuming a canonical ordering on molecular graphs or using complex equivariant architectures. 
Compared with the sequence-based models, Table \ref{tab:seq} shows that MING significantly outperforms both Character- and Grammar-VAE on QM9. These string-based models hardly learn valency rules between atom and bond types to compose valid, unique molecules. For QM9, models with high validity scores often generate lower novel molecules. This is due to the fact that the dataset enumerates all possible small valid molecules composed of only four atoms. However, this problem can be alleviated when training on larger molecules such as ZINC250k, where we observe that MING consistently achieves near-perfect novelty scores when its validity exceeds 99$\%$ as shown in Figure \ref{fig:dstep}. 
\begin{figure}[t]
\centerline{
\includegraphics[width=.9\columnwidth]{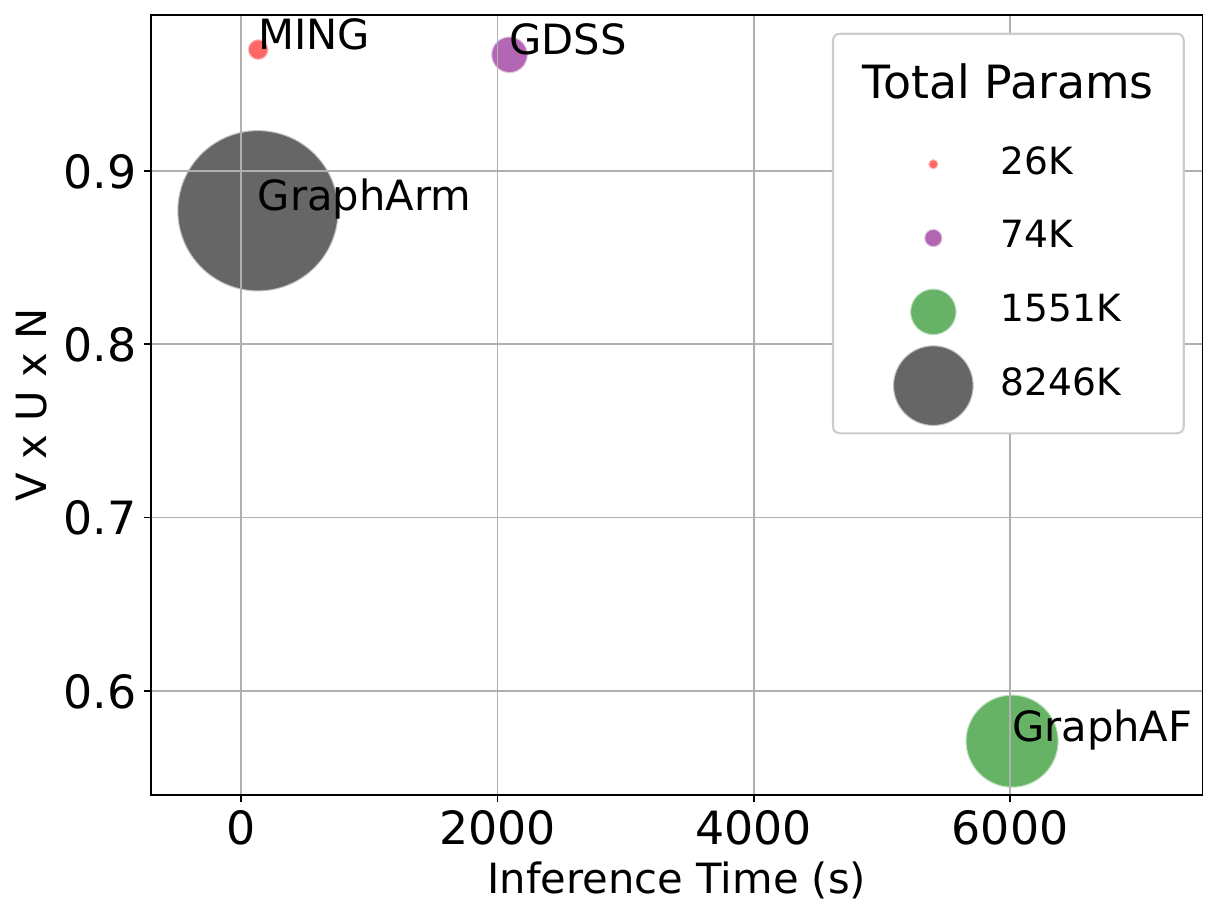}}
\caption{Benchmarking model efficiency on ZINC250k}
\label{fig:effi}
\end{figure}
\paragraph{Distribution-based Distance Metrics} Table \ref{tab:graph} shows that MING generates molecules that closely match the test distributions in terms of chemical and biological properties, measured by the FCD metric, performing comparatively to other generative models. On the larger ZINC250K dataset, MING demonstrates a clear advantage, when improving the FCD metric $2.5 \times$ compared to the best baseline, CatFlow. This highlights a promising result of learning chemical molecular properties in the function space. Furthermore, MING captures molecular graph structures more accurately, generating underlying molecular substructures—including bond and atom types—that are closer to the test distributions, as measured by the NSPDK metric. These function-based advantages are especially more evident when MING is trained on large molecule datasets.

\paragraph{Large Scale Molecular Graph Generation} We benchmark our model on the MOSES dataset, which, to the best of our knowledge, has been scaled up by very few graph generative models. Table \ref{tab:moses} presents a comparison with DiGress, a discrete diffusion framework for molecular graph generation. As observed, MING significantly enhances the validity of generated molecules while maintaining absolute novelty and a high uniqueness score. These results underscore the advantages of functional representations for molecular generation, particularly when evaluated on a challenging dataset like MOSES. Notably, we exclude the FCD score for DiGress, as their computation was based on the training set, whereas we calculate it using the test set to mitigate overfitting.

\paragraph{Speed, Memory, and Model Performance} We benchmark the model's efficiency by generating 10,000 molecules on a Titan RTX with 8 CPU cores. Figure \ref{fig:effi} visualizes the inference time, the number of model parameters, and performance, measured by the product of validity, uniqueness, and novelty VxUxN. We observe that MING shows remarkable efficiency, using $317 \times$ fewer parameters than the fastest baseline, GraphArm, while maintaining superior performance. MING is also faster than the best-performing baseline, GDSS, which applies continuous score-based models on the graph representation of molecules. Compared to the auto-regressive flow baseline GraphAF, MING significantly improves across all criteria. We do not report the model efficiency of GraphDF as this framework has a model size similar to GraphAF but $10 \times$ slower in inference speed. Additionally, DiGress and CatFlow do not open-source their model weights on ZINC250k.

\paragraph{Ablation on Diffusion Steps} We asses the MING's generation capability by ablating the different number of diffusion steps, $T \in \{30, 100, 500\}$, on ZINC250k. We train models with the same hyper-parameter set except for the diffusion steps $T$. Figure \ref{fig:dstep} presents the mean validity, uniqueness, novelty (V.U.N) results on three samplings of 10000 molecules. We observe that utilizing higher diffusion steps, $T \in \{100, 500\}$, MING can achieve over 99$\%$ validity scores. Furthermore, all experimented models consistently surpass 99.5$\%$ for uniqueness and novelty scores. By working on the function space of molecules, MING significantly speeds up the sampling process for diffusion models without causing a dramatic performance drop.

\begin{figure}[t]
\centerline{
\includegraphics[width=1.\columnwidth]{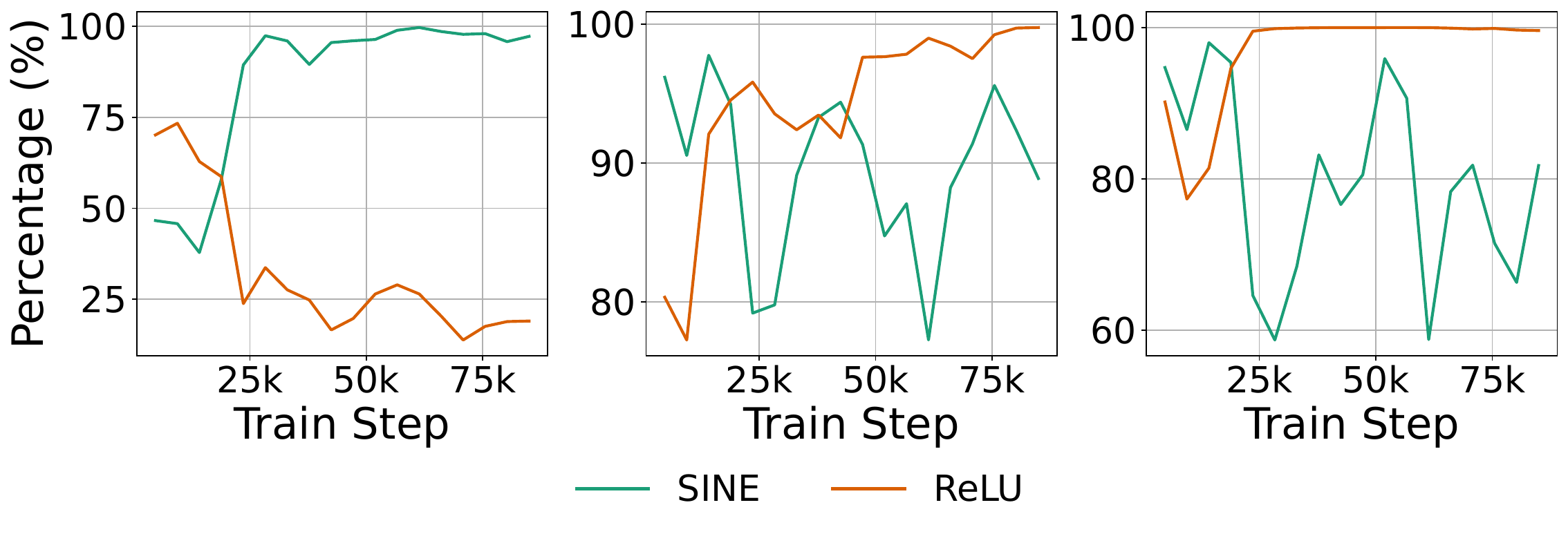}}
\caption{Validity (Left), uniqueness (Middle), and novelty (Right) results on QM9 using SINE and ReLU activation functions.}
\label{fig:act_vun}
\end{figure}

\paragraph{Ablation on non-Linear Activation Functions} We study the impact of synthesizer's activation function, $\sigma$, on model performance. We train two models with the same hyper-parameter set, applying the Sinusoidal activation (SINE) on one model and the activation ReLU on the other. Figure \ref{fig:act_vun} shows the V.U.N results on 10000 sampled molecules during the training process. We remark that the synthesizer with ReLU is incapable of modeling molecular signals on their topological space, leading to low validity scores for the generated molecules. On the other hand, the synthesizer using the activation SINE learns the function representation of molecules better, mapping from topological inputs to molecular signals. These findings are consistent with \cite{sitzmann2019scene}, where they first experimented with the Sinusoidal activation for representing complex data signals on regular domains.

\section{CONCLUSION \& LIMITATIONS}

This paper introduces MING, a generative model that learns molecule distributions in function space by optimizing an expectation-maximization denoising objective. To our knowledge, MING is the first function-based generative model capable of generating both bond and atom types of molecules. Our experiments on molecule datasets demonstrate several advantages of leveraging functional representations for molecule generation. These include faster inference speed, smaller model size, and improved performance in capturing molecule distributions as measured by statistics-based and distribution-based metrics.
Unlike graph-based models, MING avoids the complexities of graph-permutation symmetry, streamlining model architecture design. This approach could align with recent research exploring the potential of non-equivariant molecular generative models.
A limitation of the current work is its reliance on resampling known valid coordinates from training datasets when generating molecular signals through the reverse diffusion process. This constraint may, to some extent, restrict the model's ability to generate diverse molecular structures beyond data distributions, which could be valuable for certain molecular applications. In future work, we will explore learning prior distributions of valid coordinates to enhance the model's flexibility.
We also aim to extend the framework to address broader molecule-related problems in the function space, such as molecule conformation and entire 3D physical structure generation.


\subsubsection*{Acknowledgements}
We acknowledge the financial support of the Swiss National Science Foundation within the LegoMol project (grant no. 207428). The computations were performed at the University of Geneva on Baobab and Yggdrasil HPC clusters. We also thank the anonymous reviewers at AISTATS for their insightful feedback and suggestions.

\bibliography{aistats2025/ming}






\section*{Checklist}



 \begin{enumerate}

 \item For all models and algorithms presented, check if you include:
 \begin{enumerate}
   \item A clear description of the mathematical setting, assumptions, algorithm, and/or model. [Yes]
   \item An analysis of the properties and complexity (time, space, sample size) of any algorithm. [Yes]
   \item (Optional) Anonymized source code, with specification of all dependencies, including external libraries. [Yes]
 \end{enumerate}

 \item For any theoretical claim, check if you include:
 \begin{enumerate}
   \item Statements of the full set of assumptions of all theoretical results. [Yes]
   \item Complete proofs of all theoretical results. [Yes]
   \item Clear explanations of any assumptions. [Yes]     
 \end{enumerate}

 \item For all figures and tables that present empirical results, check if you include:
 \begin{enumerate}
   \item The code, data, and instructions needed to reproduce the main experimental results (either in the supplemental material or as a URL). [Yes]
   \item All the training details (e.g., data splits, hyperparameters, how they were chosen). [Yes]
         \item A clear definition of the specific measure or statistics and error bars (e.g., with respect to the random seed after running experiments multiple times). [Yes]
         \item A description of the computing infrastructure used. (e.g., type of GPUs, internal cluster, or cloud provider). [Yes]
 \end{enumerate}

 \item If you are using existing assets (e.g., code, data, models) or curating/releasing new assets, check if you include:
 \begin{enumerate}
   \item Citations of the creator If your work uses existing assets. [Yes]
   \item The license information of the assets, if applicable. [Not Applicable]
   \item New assets either in the supplemental material or as a URL, if applicable. [Not Applicable]
   \item Information about consent from data providers/curators. [Not Applicable]
   \item Discussion of sensible content if applicable, e.g., personally identifiable information or offensive content. [Not Applicable]
 \end{enumerate}

 \item If you used crowdsourcing or conducted research with human subjects, check if you include:
 \begin{enumerate}
   \item The full text of instructions given to participants and screenshots. [Not Applicable]
   \item Descriptions of potential participant risks, with links to Institutional Review Board (IRB) approvals if applicable. [Not Applicable]
   \item The estimated hourly wage paid to participants and the total amount spent on participant compensation. [Not Applicable]
 \end{enumerate}

 \end{enumerate}

\newpage
\appendix
\onecolumn


\section{MING DETAILED MATHEMATICAL DERIVATIONS}

We present denoising diffusion probabilistic models applied to the function representation of molecules. In this framework, we represent each molecule function as a set of function evaluations over its input domain, formally defined as:
\begin{align}
    f_0 \triangleq \{ s_0 \triangleq (\phi_0, z_0, y_0) | \phi_0 \in \mathcal{T}, z \in \mathcal{Z}, y \in \mathcal{Y}\}
\end{align}
where $\mathcal{T}$ is the molecule input domain,  $\mathcal{Y}$ is the corresponding molecular signals on $\mathcal{T}$, $\mathcal{Z}$ is an unknown latent manifold, and $s_0$ is the function evaluation on the topology input $\phi_0$. Following the approach from \cite{sohl2015deep,ho2020denoising}, we want to learn a model $\theta$ that maximizes the likelihood given an initial function evaluation $s_0$. This can be formulated as:
\begin{align}
    p_{\theta}(s_0) = \int p_{\theta}(s_{0:T})ds_{1:T}
\end{align}

where $s_{1:T}$ represents a series of noised function evaluations during the diffusion process. However, directly evaluating the likelihood is intractable, we sort to the relative evaluation between the forward and reverse process, averaged over the forward process \citep{sohl2015deep}:
\begin{align}
    p_{\theta}(s_0) &= \int p_{\theta}(s_{0:T})ds_{1:T} \nonumber \\
    &= \int p_{\theta}(s_{0:T}) \frac{q(s_{1:T} | s_0)}{q(s_{1:T} | s_0)} ds_{1:T} \nonumber \\
    &= \int p_{\theta}(s_{0:T}) \frac{q(s_{1:T} | s_0)}{q(s_{1:T} | s_0)} ds_{1:T} \nonumber \\
    &= \int p_{\theta}(s_{T}) q(s_{1:T} | s_0) \prod \frac{p_{\theta}(s_{t-1}|s_{t})}{q(s_{t} | s_{t-1})}ds_{1:T} \nonumber \\
\end{align}
where $q(s_{1:T} | s_0)=\prod q(s_t| s_{t-1})$ defines a Markovian forward trajectory. Training involves minimizing the lower bound of the negative log likelihood:
\begin{align}
    \mathbb{E}_{q(s_{1:T} | s_0)} \left[ - \log p_{\theta} (s_0)\right] \leq  \mathbb{E}_{q(s_{1:T} | s_0)} \left[ - \log \frac{p_{\theta}(s_{0:T})}{q(s_{1:T} | s_0 )}\right] \triangleq \mathcal{L}_{\theta}
\end{align}

This can further be decomposed into a sum over diffusion steps:
\begin{align}
    \mathcal{L}_{\theta} = \mathbb{E}_{q(s_{1:T} | s_0 )} \left[\underbrace{-\log p_{\theta}(s_0|s_1)}_{\mathcal{L}_0} + \sum_{t=2}^T\underbrace{ D_{KL}\left(q(s_{t-1}|s_t, s_0) \| p_{\theta}(s_{t-1}|s_{t})\right)}_{\mathcal{L}_{t-1}} + \underbrace{D_{KL}\left(q(s_{T}| s_0) \| p_{\theta}(s_T)\right)}_{\mathcal{L}_T}\right]
\end{align}

where $q(s_{t-1}|s_t, s_0) $ and $ p_{\theta}(s_{t-1}|s_{t})$ are the forward conditional and denoising model posteriors which we analyse in the functional setting as follows.

\subsection{Conditional Forward Posterior}
\label{proof:cfp}

Substituting the triplet-based function evaluation to the conditional forward posterior we have:
\begin{align}
    q(s_{t-1}|s_{t}, s_{0}) &= q (\phi_{t-1}, z_{t-1}, y_{t-1} | \phi_{t}, z_{t}, y_{t}, \phi_{0}, z_{0}, y_{0}) \nonumber \\
    &= q (\phi_{0}, z_{t-1}, y_{t-1} | \phi_{0}, z_{t}, y_{t}, \phi_{0}, z_{0}, y_{0}) && \text{(topology-input invariance.)} \nonumber \\
    &= q (z_{t-1}, y_{t-1} | z_{t}, y_{t}, \phi_{0}, z_{0}, y_{0}) \nonumber \\
    &= q (z_{t-1} | z_{t}, y_{t}, y_{t-1}, \phi_{0}, z_{0}, y_{0})  q ( y_{t-1} | z_{t}, y_{t}, \phi_{0}, z_{0}, y_{0}) \nonumber \\
    &= q (z_{t-1} | y_{t-1}, \phi_{0}) q ( y_{t-1} | z_{t}, y_{t}, \phi_{0}, z_{0}, y_{0}) && \text{(latent temporal conditional independence.)} \nonumber \\
    &= q (z_{t-1} | y_{t-1}, \phi_{0}) q ( y_{t-1} | \underbrace{\phi_{0}, z_{t}}_{y_t}, y_{t}, \underbrace{\phi_{0}, z_{0}}_{y_0}, y_{0}) \nonumber \\
    &= q (z_{t-1} | y_{t-1}, \phi_{0}) q ( y_{t-1} | y_{t},  y_{0}) 
\end{align} 

\subsection{Denoising Model Posterior}

In a similar way we derive the denoising model posterior: 
\begin{align}
p_{\theta}(s_{t-1} | s_{t}) &= p_{\theta}(z_{t-1}, \phi_{t-1}, y_{t-1} | z_{t}, \phi_{t}, y_{t}) \nonumber \\
&= p_{\theta}(z_{t-1}, \phi_{0}, y_{t-1} | z_{t}, \phi_{0}, y_{t}) && \text{(topology-input invariance.)} \nonumber \\
&= p_{\theta}(z_{t-1}, y_{t-1} | z_{t}, \phi_{0}, y_{t}) \nonumber\\
&= p_{\theta}(z_{t-1} | z_{t}, \phi_{0}, y_{t}, y_{t-1})p_{\theta}(y_{t-1} | z_{t}, \phi_0, y_t) \nonumber\\
&= p_{\theta}(z_{t-1} | \phi_{0}, y_{t-1}) p_{\theta}(y_{t-1} | \underbrace{z_{t}, \phi_0}_{\hat{y}_0}, y_t) && \text{(latent temporal conditional independence.)}
\end{align}

\subsection{Proof of Proposition \ref{prop:L}}
\label{proof:p1}

In order to prove  Proposition \ref{prop:L}, we replace the functional posterior forms above to the KL term $\mathcal{L}_{t-1}$:
\begin{align}
\mathcal{L}_{t-1} &= D_{KL}(q(s_{t-1} | s_{t}, s_{0}) \| p_{\theta}(s_{t-1} | s_{t})) \nonumber \\
&= \int q(s_{t-1} | s_{t}, s_{0}) \log\frac{q(s_{t-1} | s_{t}, s_{0})}{p_{\theta}(s_{t-1} | s_{t})} ds_{t-1} \nonumber \\
&= \int\int q (z_{t-1} | y_{t-1}, \phi_{0}) q(y_{t-1} | y_{t}, y_{0}) \log\frac{q (z_{t-1} | y_{t-1}, \phi_{0}) q(y_{t-1} | y_{t}, y_{0})}{p_{\theta}(z_{t-1} | \phi_{0}, y_{t-1}) p_{\theta}(y_{t-1} | z_{t}, \phi_0, y_t)} dz_{t-1}dy_{t-1} \nonumber \\
&= \int\int q (z_{t-1} | y_{t-1}, \phi_{0}) q(y_{t-1} | y_{t}, y_{0}) \log \frac{q (z_{t-1} | y_{t-1}, \phi_{0})}{p_{\theta}(z_{t-1} | \phi_{0}, y_{t-1})}dz_{t-1}dy_{t-1} \nonumber \\ 
&\qquad\qquad\qquad  + \int\int q (z_{t-1} | y_{t-1}, \phi_{0}) q(y_{t-1} | y_{t}, y_{0}) \log \frac{q(y_{t-1} | y_{t}, y_{0})}{p_{\theta}(y_{t-1} | z_{t}, \phi_0, y_t)}dz_{t-1}dy_{t-1} \nonumber \\ 
&= \int\int q (z_{t-1} | y_{t-1}, \phi_{0}) \log \frac{q (z_{t-1} | y_{t-1}, \phi_{0})}{p_{\theta}(z_{t-1} | \phi_{0}, y_{t-1})}dz_{t-1}\;q(y_{t-1} | y_{t}, y_{0})dy_{t-1}  \nonumber \\ 
&\qquad\qquad\qquad  + \int q(y_{t-1} | y_{t}, y_{0}) \log \frac{q(y_{t-1} | y_{t}, y_{0})}{p_{\theta}(y_{t-1} | z_{t}, \phi_0, y_t)}dy_{t-1} \underbrace{\int q (z_{t-1} | y_{t-1}, \phi_{0})dz_{t-1}}_{constant}\nonumber \\ 
&= \mathbb{E}_{q(y_{t-1} | y_t, y_0)}\left(D_{KL}\left(q(z_{t-1}|y_{t-1}, \phi_0) \| p_{\theta} (z_{t-1} | y_{t-1}, \phi_0)\right)\right)+ D_{KL} \left(q(y_{t-1}| y_t, y_0) \| p_{\theta}(y_{t-1}|z_t, \phi_0, y_t)\right) \nonumber \\ 
&= \mathcal{L}_{t-1}^{\mathcal{Z}} + \mathcal{L}_{t-1}^{\mathcal{Y}}
\end{align}

We have the KL term $\mathcal{L}_{t-1}$ can be directly optimized on the latent input space, $\mathcal{Z}$, and signal output space, $\mathcal{Y}$.

\subsection{Proof of Proposition \ref{prop:Z}}
\label{proof:p2}

Given a latent model $p_{\psi}\left(z, \phi, y\right) = p(z, \phi, y| \psi)$ where $\psi: \phi \times z \rightarrow y$ models the relation between the topology input $\phi$, latent variable $z$, and signal output $y$. For an observed pair of topology input and signal $(\phi, y)$, we look for $z$ to minimize an optimization problem:
\begin{align}
    z = \argmin_{z} \|\psi(\phi, z) - y\|^2
\end{align}

 Since $\psi$ is an INR network, which is differentiable and deterministic, the loss gradient w.r.t $z$ is thus uniquely determined for each value of $z$. By using gradient descent optimization to the minimization problem, at the step $k+1$ we have $z_{t+1}$ is uniquely defined:
\begin{align}
    z_{k+1} = z_{k} - \nabla_{z_k}\|\psi(\phi, z_k) - y\|^2
\end{align} 

We apply the latent model $p(z, \phi, y| \psi)$ to the conditional forward posterior $q(z_{t-1}| y_{t-1}, \phi_0)$ and the denoising model posterior $p_{\theta}(z_{t-1}| y_{t-1}, \phi_0)$. Since these posteriors are conditioned on the same topology input and signal output $(\phi_0, y_{t-1} \sim q(y_{t-1} | y_t, y_0))$, the latent model results in the same latent-input representation $z_{t-1}$ for both posteriors.

\section{EXPERIMENTAL RESULTS}

\subsection{Additional Experiments}

\subsubsection{Ablation on Topological Input Dimensions}

\begin{figure*}[th]
\centerline{
\includegraphics[width=1.\textwidth]{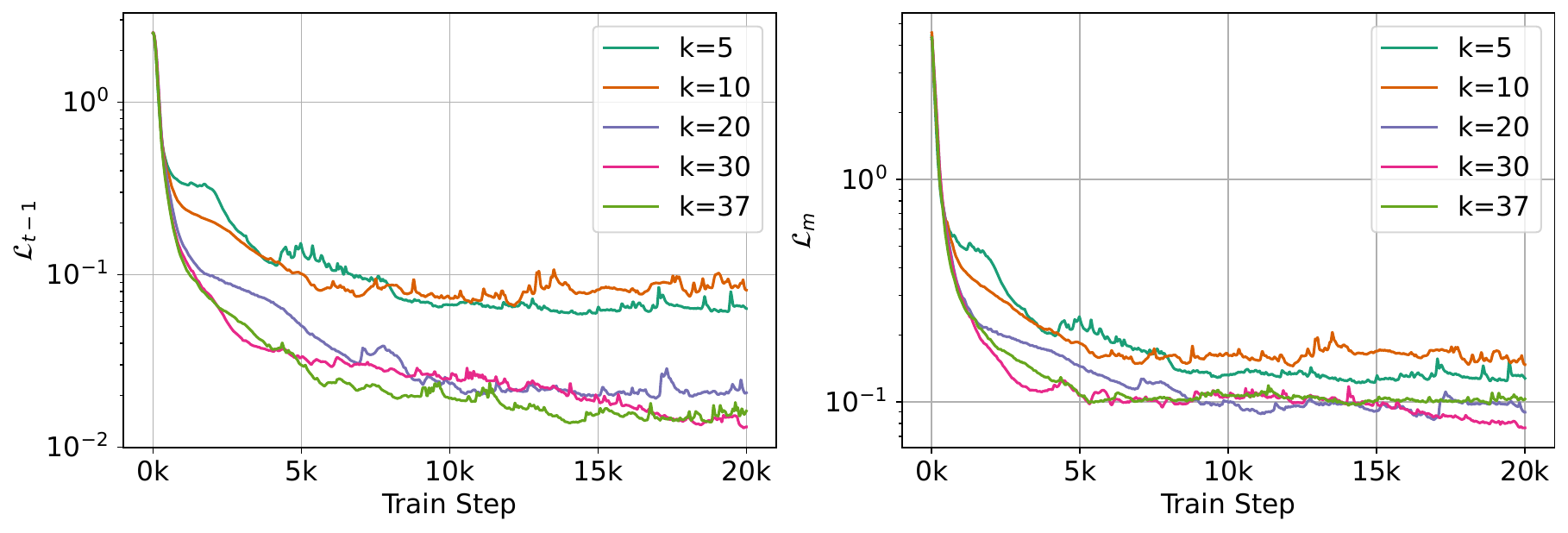}}
\caption{Ablation study on topological input dimensions. The denoising objective loss (\textit{left}), $\mathcal{L}_{t-1} (\theta)$, and maximization objective loss (\textit{right}), $\mathcal{L}_{m}(\theta, \psi)$, for different topological input dimensions, $k$.}
\label{fig:topo}
\end{figure*}

We investigate the impact of structural changes in the topology input, $\phi$, on model denoising efficiency. We experiment on ZINC250k as the dataset features a high topological dimension, with $k_{max}=38$. Figure \ref{fig:topo} shows the denoising loss, $\mathcal{L}_{t-1} (\theta)$, and maximization loss, $\mathcal{L}_{m}(\theta, \psi)$, across different topology-input sizes. We observe that MING can achieve more effective denoising with higher-dimensional inputs. However, when experimenting nearly the full topological dimension of the data, $k=37$, the model's denoising ability slightly declines. Utilizing an suitable dimension allows MING to generalize well across varying molecule's topological input spaces.

\subsubsection{Ablation on Inner Optimization Steps}

\begin{figure*}[th]
\centerline{
\includegraphics[width=1.\textwidth]{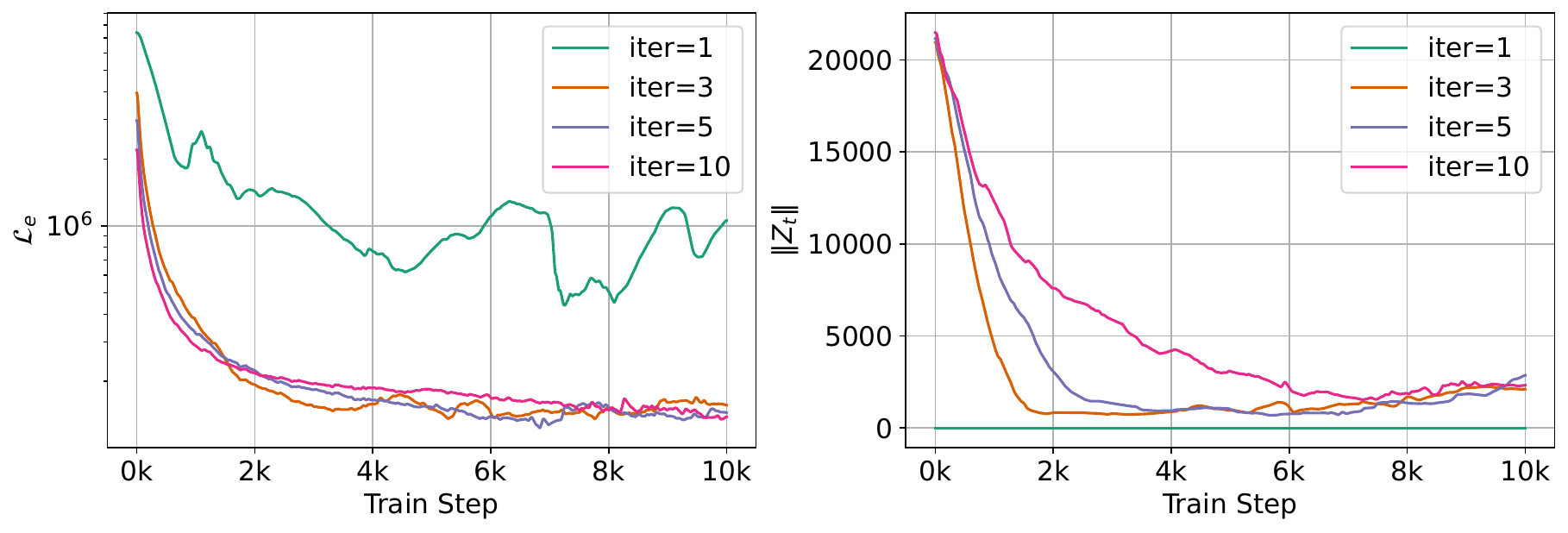}}
\caption{Ablation study on inner optimization steps.  The sum of  expectation objective losses (\textit{left}), $\mathcal{L}_e(z_t)$, and z-norm regularisation (\textit{right}) for different numbers of inner optimisation steps.}
\label{fig:instep}
\end{figure*}

We ablate the effect of inner optimization steps on the expectation objective loss, $\mathcal{L}_e(z_t)$. As depicted in Figure \ref{fig:instep}, MING significantly reduces the expectation loss and stabilizes latent input learning with a higher number of optimisation steps. Moreover, choosing an appropriate iterations is essential for balancing the tradeoff between MING's inference speed and performance.

\subsection{Main Experiments}
\label{sec:exp}

\subsubsection{Hyperparameters}

\begin{table*}[th]
\caption{MING hyperparameters.}
\begin{center}
\begin{small}
\begin{tabular}{ll ccc}
\toprule
 & Hyperparam & QM9 & ZINC250k & MOSES\\
\midrule
\multirow{5}{*}{$\phi, \psi$} & topology-input dimension $d$ & 7 & 30 & 25\\
 & hidden dimension & 256 & 64 & 128\\
 & latent dimension $k$ & 64 & 8 & 32\\
 & signal dimension $f$ & 8 & 13 & 12\\
 & number layers & 8 & 3 & 5\\
\midrule
\multirow{3}{*}{SDE} & $\beta_{\min}$ & 0.0001 & 0.0001 & 0.0001\\
 & $\beta_{\max}$ & 0.02 & 0.02 & 0.02\\
 & diffusion steps $T$ & 100 & 30 & 100\\
\midrule
\multirow{4}{*}{MING} & $\mathcal{L}_e$ optimizer  & Gradient descent & Gradient descent & Gradient descent \\
&  $\mathcal{L}_m$ optimizer  & Adam & Adam & Adam\\
&  $\mathcal{L}_e$ learning rate  & 0.1 & 0.1 & 0.1\\
&  $\mathcal{L}_m$ learning rate  & 0.0001 & 0.001 & 0.001\\
&  $\mathcal{L}_e$ number optimisation steps  & 3 & 3 & 3\\
& batch size  & 256  & 256 & 256\\
& number of epochs  & 500 & 500 & 500\\
\bottomrule
\end{tabular}
\end{small}
\end{center}
\label{table:glad-hyper}
\end{table*}

\subsubsection{Detailed Results}

\begin{table*}[th]
\caption{Molecule graph statistics.}
\label{tab:data}
\begin{center}
\begin{small}
\begin{tabular}{lcccc}
\toprule
 & Graphs & Nodes & Node Types & Edge Types \\
\midrule
\textsc{QM9} & $133885$ & $1  \leqslant |X| \leqslant 9$ & $4$ & $3$ \\
\textsc{ZINC250k} & $249455$ & $6  \leqslant |X| \leqslant 38$  & $9$ & $3$\\
\textsc{MOSES} & $1936962$ & $8  \leqslant |X| \leqslant 27$  & $7$ & $4$\\
\bottomrule
\end{tabular}
\end{small}
\end{center}
\end{table*}

\begin{table*}[th]
\caption{MING's detailed results, including the mean and standard deviations for three samplings.}
\label{tab:mol-detail}
\begin{center}
\begin{small}
\begin{tabular}{lccccc} 
\toprule
 & Val. $\uparrow$ &  Uni. $\uparrow$ & Nov. $\uparrow$ & NSPDK $\downarrow$ & FCD $\downarrow$ \\ 
\midrule
\textsc{$\text{ZINC250k}$} & 97.19 $\pm$ 0.09 & 99.92 $\pm$ 0.03 & 99.84 $\pm$ 0.06 & 0.005 $\pm$ 0.000 & 4.8 $\pm$ 0.02 \\
\textsc{$\text{QM9}$} & 98.23 $\pm$ 0.09 & 96.83 $\pm$ 0.18 & 71.59 $\pm$ 0.21 & 0.002 $\pm$ 0.000 & 1.17 $\pm$ 0.00 \\
\textsc{$\text{MOSES}$} & 97.23 $\pm$ 0.10 & 99.88 $\pm$ 0.01 & 100 $\pm$ 0 & - & 15.03 $\pm$ 0.02 \\
\bottomrule
\end{tabular}
\end{small}
\end{center}
\end{table*}

\subsubsection{Visualizations}
\begin{figure*}[th]
\centerline{
\includegraphics[width=1.\textwidth]{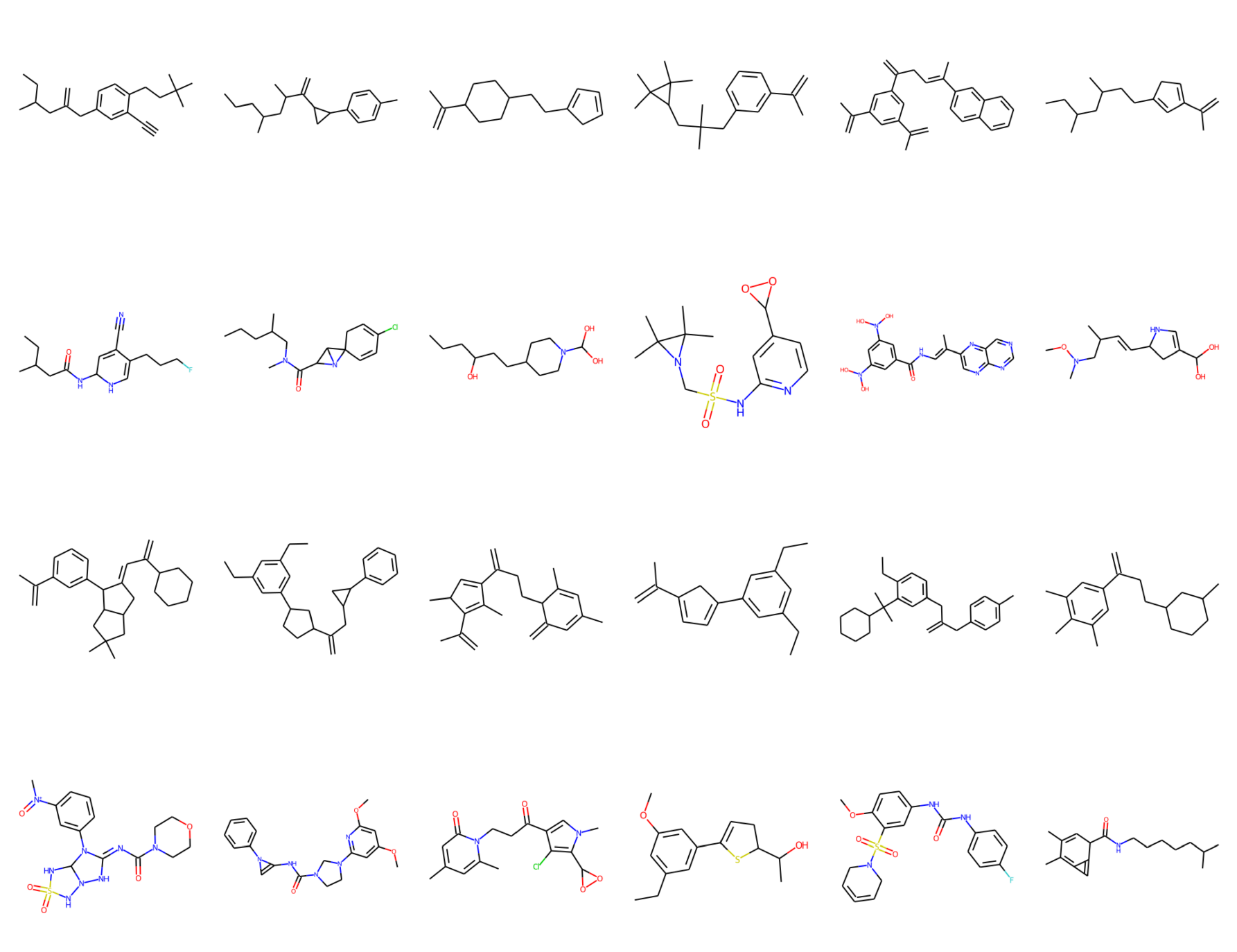}}
\caption{Visualizations of the topology input samples (top) and their corresponding generated molecules (bottom) from ZINC250k.}
\label{fig:vizmol}
\end{figure*}

\end{document}